# Robust Optimization through Neuroevolution

Paolo Pagliuca and Stefano Nolfi
Institute of Cognitive Sciences and Technologies
National Research Council (CNR)
Via S. Martino della Battaglia, 44
00185 Roma, Italia

**Abstract**

*We propose a method for evolving solutions that are robust with respect to variations of the environmental conditions (i.e. that can operate effectively in new conditions immediately, without the need to adapt to variations). The obtained results show how the method proposed is effective and computational tractable. It permits to improve performance on an extended version of the double-pole balancing problem, to outperform the best available human-designed controllers on a car racing problem, and to generate rather effective solutions for a swarm robotic problem. The comparison of different algorithms indicates that the CMA-ES and xNES methods, that operate by optimizing a distribution of parameters, represent the best options for the evolution of robust neural network controllers.*

## 1. Introduction

Many real-world optimization problems require the development of systems capable of operating in environmental conditions that change stochastically and/or dynamically over time (Jin and Branke, 2005; Cruz, Gonzalez and Pelta, 2011; Branke, 2012). The physical world is full of uncertainties and almost no parameter, dimension, or property tends to remain exactly constant. Machines may break down or wear out slowly, raw material is of changing quality, the physical conditions of the environment change continuously as a result of meteorological phenomena and/or as a result of the activities of other systems, new jobs might need to be integrated with activities already scheduled.

To operate effectively in variable environmental conditions systems should be able to adapt dynamically to environmental variations and/or should be robust, i.e. should be able to operate as satisfactory as possible in varied environmental conditions immediately, without the need to further adapt. In certain conditions, e.g. when the environment changes too quickly, when the environment cannot be monitored closely enough, or when the cost associated to the time required to adapt to the new environmental conditions is too high, robustness represents the only possible way to handle the effect of environmental variations (Branke, 2012, chapter 8).

In this paper we investigate the use of artificial evolution for robust design optimization (Beyer and Sendhoff, 2007), namely for the synthesis of solutions capable to operate satisfactorily in varied environmental conditions immediately. This is investigated in the context of neuro-controlled agents evolved in varying environmental conditions. We decided to use agents situated in an external environment since this permits to study the effect of environmental and agent/environmental variations. Moreover, we decided to use neuro-controlled agents to exploit the ability of neural network to regulate their output on the basis of their input and to generalize, i.e. to respond to new inputs with outputs that are similar to those produced for similar inputs. The possibility to regulate the output on the basis of input units that encode the state of the environment permits to display multiple behaviors and to select the behavior that is adapted to the current environmental circumstances (Carvalho and Nolfi, 2017). The tendency to respond to new inputs with outputs that are similar to those produced for similar inputs increases the chance to master appropriately new environmental conditions without the need to adapt to them.

After reviewing the previous works carried out in this area, we introduce a method that can be used to evolve robust solution that specify: how the fitness of candidate solutions can be evaluated, how the variation of the environmental conditions during the course of the evolutionary process can be controlled to improve the quality of the evolved solutions, how the best solution of a run can be identified, and which are the characteristics that the evolutionary algorithm should have to operate effectively in varying environmental conditions.

The method is validated on three qualitatively different problems: (1) an extended version of the double-pole balancing problem, (2) a car-racing problem, and (3) a swarm robotic problem.

Results are collected by using a standard $(\mu + \mu)$ evolutionary strategy (Rechenberg, 1973; Pagliuca, Milano, and Nolfi, 2018) and four state-of-the-art evolutionary algorithms: (1) the covariance matrix adaptation evolution strategy (CMA-ES, see Hansen and Ostermeier, 2001), (2) the exponential natural evolutionary strategy (xNES, see Wierstra, Schaul, Peters and Schmidhuber, 2008), (3) the separable natural evolutionary strategy (sNES, see Wierstra, Schaul, Peters and Schmidhuber, 2008), and (4) the neuroevolution of augmenting topologies (NEAT, Stanley and Miikkulainen, 2002).

The obtained results show how solutions displaying a high level of robustness can be obtained by evaluating candidate solutions in a relatively small number of different environmental conditions. Moreover, the results show that the CMA-ES, xNES and sNES algorithms, that operate by optimizing distribution of parameters, outperform alternative algorithms, that operate by optimizing specific combination of parameters.

**2. Related research**

Uncertainties in evolutionary system embedded in an external environment can occur as a consequence of variations affecting: (i) the internal environment of the candidate solution, and/or (ii) the external environment, and/or (iii) the evaluation of the objective function (see Figure 1; Branke, 2002; Beyer and Sendhoff, 2007). "The goal of robust optimization therefore is twofold: (1) to find optimal solution despite uncertainties and noise in the optimization model, and (2) to find optimal solutions that are robust with respect to uncertainties and noise, and therewith useful in practice." (Kruisselbrink, 2012; pp. 213).

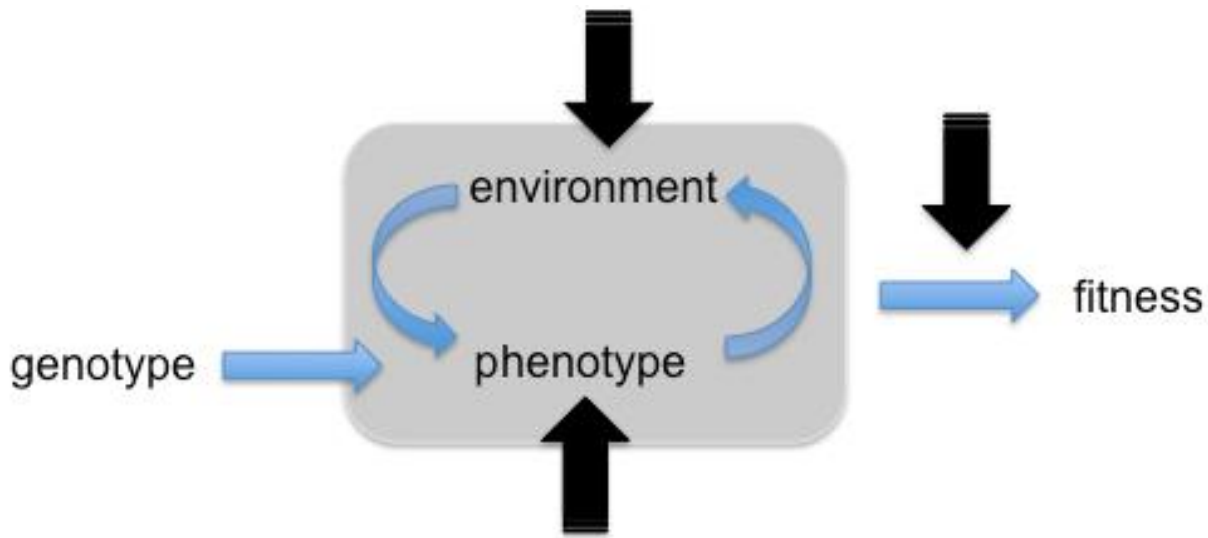

Figure 1. Schematization of the components of an evolutionary system that can be subjected to variations. The genotype determines the characteristics of the phenotype (solution). The continuous interactions between the phenotype and the environment produce a behavior (represented by the grey rectangular area). The fitness indicates the efficacy of the behavior. As indicated by the black arrows, variations might affect the phenotype (i.e. the internal environment), and/or the environment (i.e. the external environment or the relation between the agent and the environment), and/or the fitness evaluation. Adapted from Figure 8.3 (Branke, 2012).

The most general approach to evolve robust systems consists in evaluating candidate solutions multiple times in varying environmental conditions, i.e. carrying out multiple evaluation episodes characterized by different environmental conditions. Different type of robustness can be promoted through the usage of fitness functions that reward for: (i) the maximization of the expected fitness, i.e. the average fitness obtained in all possible environmental conditions (Branke, 2012), (ii) the maximization of the worst-case fitness (McIlhagga, Husbands and Ives, 1996), (iii) the minimization of the variations of fitness in varying conditions (Branke, 2012), (iv) the maximum of the amount of tolerable disturbances (Deb, Gupta, Daum, Branke, Mall and Padmanabhan, 2009), (v) the maximization of the ratio between the standard deviation of the fitness and the amount of environmental variation (Jin and Sendhoff, 2003). When the fitness function returns a single value a standard evolutionary algorithm is used. Instead, when the fitness function returns multiple values, e.g. the expected fitness and the fitness variance, the usage of a multi-objective evolutionary algorithm can be beneficial (Deb and Gupta, 2006). In our experiments we will consider the optimization of the expected fitness only.

Overall the evolution of robust solutions requires the determination of the following aspects: (i) how fitness is estimated, (ii) how the environmental conditions vary along the evolutionary process (at least in the case in which the experimenter has the possibility to manipulate them), (iii) how the best solution of a run is identified, and (iv) which evolutionary algorithm is used.

Fitness can be measured by evaluating candidate solutions in all possible environmental conditions. An exhaustive evaluation of this type, however, is possible only when the number of environmental parameters that can vary and the number of states that they can assume is small. In the other cases, fitness should be estimated by evaluating candidate solutions on a subset of all the possible environmental conditions. The greater the number of evaluation episodes carried out under different environmental conditions is, the greater the precision of the estimation is. On the other hand, the higher the number of evaluation episodes is, the higher the computational cost is. This implies that one should identify a good balance between accuracy and cost. The problem of measuring fitness in variable environmental conditions should not be confused with the problem of estimating fitness in the presence of measuring errors. Indeed, in the latter case, each candidate solution has a single fitness level that is affected by measuring errors. In the former case, instead, each candidate solution has N potentially different fitness levels in N different corresponding environmental conditions. This implies that the techniques that can be used to minimize the number of evaluation episodes while ensuring a sufficient precision of the fitness estimation in the case of noisy fitness functions cannot be applied directly to the case of variable environmental conditions. Examples of such techniques include: (i) the usage of statistical tests to identify the minimal number of evaluations that can be used to differentiate with a sufficient level of precision the relative fitness of alternative candidate solutions (Stagge, 1998; Cantu-Paz, 2004; Hansen et al., 2009), (ii) the usage of a large population size that permits to average implicitly the outcome of the evaluations carried out on different but similar candidate solutions (Jin and Branke, 2005; Glenn, 2013), (iii) the utilization of a non-zero threshold for accepting offspring (Markon et al. 2001), and (iv) the re-usage of the outcome of the evaluation of similar candidate solutions (Branke, 1998; Sano, Kita, Kamihira and Yamaguchi, 2000). Some researchers manage to evolve solutions that are robust with respect to environmental variations by using a single evaluation episode conduced in randomly selected environmental conditions. For example, Salimans et. al. (2017) manage to successfully evolve neural controllers for Gym-v1 continuous control benchmarks that are robust with respect to small perturbations of their initial posture. As pointed out by Rajeswaran, Lowrey and Todorow (2017), however, this only enables to tolerate a rather limited amount of variation. Solutions robust to wider environmental variations can only be obtained by exposing candidate solutions to greater environmental variations and by carrying multiple evaluation episodes.

For what concerns the selection of the conditions in which candidate solutions are evaluated one should determine how the conditions experienced by each candidate solution are chosen and whether the conditions vary among the individuals of the same population and across generations. The simplest way to choose the conditions in which candidate solutions are evaluated consists in selecting them randomly with a given distribution (e.g. uniform or Gaussian). This can be achieved by setting the value of each aspect subjected to variation randomly within a given variation range at the beginning of each evaluation episode. Alternatively, variance reduction techniques, such as Latin hypercube sampling (LHS), can be applied (Loughlin and Ranjithan, 1999; Markon, Arnold, Back, Beielstein, and Beyer, 2001). The fitness of candidate solutions that are evaluated in easier and harder conditions tend to be over-estimate and under-estimated, respectively. This over and under-estimation effects create a noise in the fitness estimation unless the candidate solutions of each generation are evaluated on the same type of varying environmental conditions (Loughlin and Ranjithan, 1999; Markon, Arnold, Back, Beielstein, and Beyer, 2001). Varying the environmental conditions across generations should be beneficial since it reduces the probability of keep selecting candidate solutions that perform well on the current environmental conditions but perform poorly in other environmental conditions. As showed by Milano, Carvalho and Nolfi (2017), the benefit might also depend on the frequency with which environmental conditions vary across generations and can be maximized at moderate frequencies (i.e. by varying the conditions every N generations rather than every generation).

The third issue concerns how to identify the best solution of an evolutionary run. The fact that the performance depends on the environmental conditions implies that selecting the candidate solution that achieved the best fitness during the evolutionary process does not guarantee to select the best solution or at least one of the best solutions, i.e. the solution that would achieve the best average performance in all possible environmental conditions. Indeed, the solution that achieved the best fitness during the course of the evolutionary process typically corresponds to a solution that encountered favorable environmental conditions but performs much poorly in other conditions. This problem can be solved by post-evaluating evolved solutions in new environmental conditions and by selecting the candidate solution displaying the best performance during post-evaluation. To limit the computational cost required by post-evaluation, one can restrict the post-evaluation to the candidate solutions of the last generation (see for example Pagliuca, Milano and Nolfi, 2018) or to the fittest candidate solution of each generation. As we will see, the latter option is computationally more expensive but also more effective.

Finally, a fourth issue concerns the properties that the evolutionary algorithm should have to operate effectively in varying environmental conditions. A first aspect concerns the selective pressure that should be sufficiently strong to compensate the implicit reduction of the selective pressure caused by the stochasticity of the selection process. This stochasticity is caused by the fact that the environmental conditions encountered in any given generation will result easier for a subset of the current candidate solutions which will have a greater chance to be selected than the other solutions. A second aspect to consider is the need to re-evaluate the fitness of candidate solutions that remain unchanged from previous generations as a result of elitism or steady state selection. This is required to avoid the un-justified permanence over generations of candidate solutions with over-estimated fitness. Finally, a third aspect to consider is the potential advantages and drawbacks of methods that estimate the local gradient and optimize a distribution of parameters (e.g. CMA-ES, xNES, and sNES; Hansen and Ostermeier, 2001; Wierstra, Schaul, Peters and Schmidhuber, 2008) with respect to methods that do not estimate the gradient and optimize a particular vector of parameters (e.g. standard Evolutionary Strategies [Beyer and Schwefel, 2002] or NEAT [Stanley and Miikkulainen, 2002]). As pointed out by Lehman, Chen, Clune and Stanley (2018), methods that optimize a distribution of parameters tend to select solutions that are robust with respect to parameters variation. Although robustness to variations of the parameters and robustness to variations of the environmental conditions are qualitatively different, solutions that are robust to parameter variations might display an above average robustness to environmental variations (Lehner, 2010). Another potential advantages of the former methods is constituted by the fact that the incremental estimation of the gradient across generations, in the presence of variable environmental conditions, implicitly produces the estimation of a gradient averaged over variable conditions. This might drive the optimization process toward area of the search space containing robust solutions. A potential drawback, instead, is constituted by the fact that the need to estimate the gradient in varying environmental condition might reduce the accuracy of the estimation.

The problem of evolving robust solutions in varying environments present similarities with the problem of evolving solutions in environment in which the environmental conditions are static but the fitness measure is noisy (Jin and Branke, 2005). Indeed, both problems usually require the usage of multiple evaluation episodes. As pointed out above, however, the two problems also present differences. In particular, the perturbations affecting the fitness measure usually have a zero mean, are symmetric, have a regular distribution, and have effects that are independent from the characteristics of the candidate solution. On the contrary, the effects of environmental variations on the fitness have a non-zero mean, are asymmetric, have a skewed distribution, and strongly depend on the characteristic of the candidate solution (Branke, 2002).

Finally, the study of evolutionary robust optimization is related to the study evolutionary dynamic optimization (Jin and Branke, 2005; Cruz, Gonzalez and Pelta, 2011; Nguyen, Yang and Branke, 2012), namely the study of how an optimization algorithm can solve an optimization problem in a given period [$t_{begin}$, $t_{end}$] in which the underlying fitness landscape changes and in which the optimization algorithm reacts to such changes by generating new optimal solutions (adapted from definition 1 included in Nguyen, Yang and Branke, 2012). Indeed, in both cases the characteristics of the environment vary. However, the objectives of robust and dynamics optimization however differ. Robust optimization aims to develop solutions capable of operating as effectively as possible in new environmental conditions immediately without further adaptation. Dynamic optimization aims to develop solutions that are not necessarily effective after the variation of the environment but that can adapt to the new environmental conditions as readily as possible. Consequently, also methodological issues differ. For example, mechanisms that preserve population diversity can speed-up the evolution of agents capable of adapting to the new environmental conditions (Cruz, Gonzalez and Pelta, 2011; Nguyen, Yang and Branke, 2012) but are not necessarily important for the evolution of robust agents. Moreover, the formalization of a method for measuring the speed with which agents adapt to the new environmental conditions is crucial for dynamic optimization (Cruz, Gonzalez and Pelta, 2011; Nguyen, Yang and Branke, 2012) but is not relevant for the study of robust agents.

## 2. Method

In this section we describe the adaptive problems considered, the evolutionary methods tested, and the method used to evolve solutions robust to environmental variations.

### 2.1 The extended double-pole balancing problem

The pole balancing problem, introduced by Wieland (1991), consists in controlling a mobile cart with one or two poles attached through passive hinge joints on the top of the cart for the ability to keep the poles balanced (Figure 2). This problem became a commonly recognized benchmark for continuous control for the following reasons: (i) it involves fundamental aspects of agent's control (e.g., situatedness, non-linearity, temporal credit assignment [Sutton, 1984]), (ii) it is intuitive and easy to understand, and (iii) it requires a low computational cost. Indeed, it has been used to compare the performance of the great majority of the neuroevolutionary methods proposed in the literature (see for example Gomez, Schmidhuber and Miikkulainen, 1998; Stanley and Miikkulainen, 2002; Igel, 2003; Durr, Mattiussi, and Floreano, 2006; Gomez, Schmidhuber and Miikkulainen, 2008; Wierstra, Schaul, Peters and Schmidhuber, 2008; Khan, Ahmad, Khan and Miller, 2013; Pagliuca, Milano, and Nolfi, 2018).

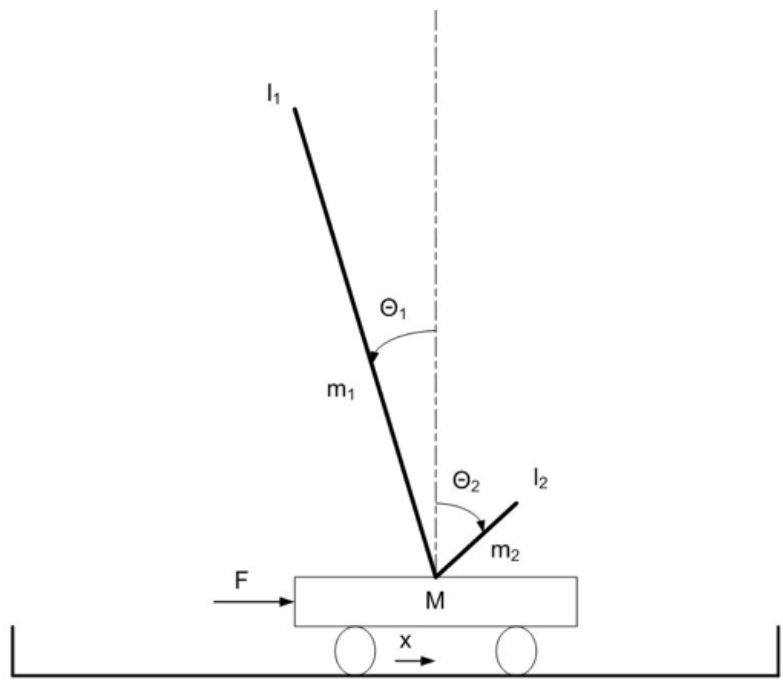

Figure 2. The double-pole balancing problem.

The cart has a mass of 1Kg. The long pole has a mass of 0.5 Kg and a length of 1.0 m. The properties of the second pole are described below. The cart can move along one dimension within a track of 4.8 m. In the non-markovian version of the problem that we consider the cart is provided with three sensors that encode the current position of the cart on the track ($x$), and the current angle of the two poles ($\theta_1$ and $\theta_2$). The motor controls the force applied to the cart along the x-axis. The goal is to maintain the angle of the poles and the position of the cart within a viable range. For a description of the equations used to calculate the dynamics of the system see Wieland (1991).

The controller of the agent is constituted by a fully connected neural network with 4 inputs (3 sensors and 1 bias unit), 10 internal neurons, and 1 motor neuron. In the case of the experiment carried with the NEAT algorithm (Stanley and Miikkulainen, 2002), the networks of the first generation only include the inputs and the output neuron but can enlarge across generations as a result of the addition of internal neurons and connections.

The inputs encode the position of the cart ($x$) and the angular position of the two poles ($\theta_1$ and $\theta_2$). The state of the $x$, $\theta_1$ and $\theta_2$ sensors are normalized in the [-0.5,0.5] m, $[-\frac{5\pi}{13.5}, \frac{5\pi}{13.5}]$ rad, and $[-\frac{5\pi}{13.5}, \frac{5\pi}{13.5}]$ rad ranges, respectively. The activation state of the motor neuron is normalized in the range [-10.0, 10.0] N and is used to set the force applied to the cart. The state of the sensors, the activation of the neural network, the force applied to the cart, and the position and velocity of the cart and of the poles are updated at a frequency of 50 Hz.

The environment in which the agent operates is stable and is constituted by a flat surface of 4.8 m. However, the initial position and velocity of the cart ($x, \dot{x}$), and the initial position and velocity of the poles ($\theta_1, \dot{\theta}_1, \theta_2, \dot{\theta}_2$), namely the relationship between the agent and the environment, can vary.

In the standard version of this problem (see for example Gomez, Schmidhuber and Miikkulainen, 2008; Stanley and Miikkulainen, 2002; Igel, 2003; Durr, Mattiussi and Floreano, 2006; Gomez, Schmidhuber and Miikkulainen, 2008; Wierstra, Schaul, Peters and Schmidhuber, 2008; Khan, Ahmad, Khan and Miller, 2013) the length/mass of the second pole is set to $\frac{1}{10}$ of that of the first pole and evolving agents are evaluated for a single episode. In most of these studies the initial state of the cart does not vary and corresponds to the following values [$x$=0.0, $\dot{x}$=0.0, $\theta_1$=0.07, $\dot{\theta}_1$=0.0, $\theta_2$=0.0, $\dot{\theta}_2$=0.0]. In the case of Khan, Ahmad, Khan and Miller (2013), instead, the initial state of the single evaluation episode is selected randomly every time within the following intervals: [$-2.4 < x < 2.4, \dot{x} = 0.0, -0.6 < \theta_1 < 0.6, \dot{\theta}_1 = 0.0, -0.6 < \theta_2 < 0.6, \dot{\theta}_2 = 0.0$] (see for example Khan, Ahmad, Khan and Miller, 2013). Clearly, however, the utilization of a single evaluation episode does not favor the selection of robust solutions, i.e. agents capable to keep the poles balanced in environmental conditions that vary freely (within limits) with respect to those parameters. The evolved agents display a certain level of generalization even in these conditions. This is due to the fact that to be successful, candidate solutions need to balance the poles also from the states that the cart and the poles assume during the course of the episode. However, the states encountered by the evolving agents during the course of their evaluation episode represent only a subset of all possible states. The utilization of a single evaluation episode, therefore, does not favor the selection of truly robust solutions, i.e. solutions capable of balancing the poles in the largest possible number of conditions, especially when the state of the agent at the beginning of the evaluation episode is kept constant.

For these reasons in the experiment reported in this article we used an extended version of the problem in which the agents are required to balance the poles during multiple evaluation episodes in which the cart and the poles assume randomly different initial states chosen within the following intervals [$-1.944 < x < 1.944$, $-1.215 < \dot{x} < 1.215$, $-0.0472 < \theta_1 < 0.0472, -0.135088 < \dot{\theta}_1 < 0.135088, -0.10472 < \theta_2 < 0.10472, -0.135088 < \dot{\theta}_2 < 0.135088$]. Moreover, to increase the complexity of the problem we set the length/mass of the second pole to $\frac{1}{2}$ of that of the first pole instead than to $\frac{1}{10}$ as in the case of the standard version of the problem. Indeed, as pointed out by Wieland (1991), the longer the second pole is, the higher the complexity of the problem becomes. The aspect that is subjected to variations, therefore, is the initial agent/environmental relation. In particular, the relative position of the cart, and the orientation and the speed of two poles.

Episodes terminate after 1000 steps or when the angular position of one of the two poles exceeded the range [$-\frac{\pi}{5}, \frac{\pi}{5}$] rad or the position of the cart exceed the range [-2.4, 2.4] m. The fitness of the agent corresponds to the fraction of time steps in which the agent maintains the cart and the poles within the allowed position and orientation ranges averaged over multiple evaluation episodes.

**2.2 The car-racing problem**

The design of car controllers for The Open Racing Car Simulator (TORCS, Espie et al., 2005) constitutes another widely used benchmark for continuous control. TORCS (torcs.sourceforge.net) is a state-of-the-art open source car-racing simulator. It combines the features of an advanced commercial simulator with those of a fully customizable research environment. It includes a rather sophisticated physics engine, which takes into account many aspects of the racing car (e.g. collisions, traction, aerodynamics, fuel consumption, etc.) and allows the users to develop their own car controllers as separate C++ modules, which can be easily compiled and added to the game. At each control step, a bot can access the current game state, which includes information about the car and the track, and can control the car using the gas/brake pedals, the gear stick, and steering wheel. The game distribution includes many programmed bots, which can be easily customized or extended to build new bots. TORCS users developed several bots, which often compete in international competitions. Thanks to these features TORCS has quickly become the de facto standard in the computational intelligence research (Loiacono, 2012).

Machine learning methods applied to this domain usually focus on the problem of identifying an optimal racing line, i.e. a trajectory that the car should follow to minimize the lap-time on a given track with a given car (Lecchi, 2009). The low-level control necessary to produces such target trajectory is then realized by a human-designed controller. As in the case of Cardamone, Loiacono and Lanzi (2010), instead, we evolve neural network controllers that learn to drive from scratch and that determine the desired state of the actuators of the car on the basis of the state of the game that can be perceived from the car-driver point of view.

The neural network controller is provided with 12 sensory neurons that encode, respectively: (i) the left and right offset of the car with respect to the left and right track edge, (ii) the angular offset between the direction of the car and the orientation of the current track segment, (iii) the current car speed, (iv) the class of the current segment, i.e. 00, 01, and 10 for straight, right curves, and left curves, respectively (v) the left or right angle curvature of the current track segment, (vi) the class of the next segment, (vii) the left or right curvature of the next track segment. All values are normalized in the range [0.0, 1.0]. In the case of the left and right offset of the car, values greater than 0.8 correspond to situations in which the barycenter of the car is outside the track. For a discussion of the importance of perceiving the curvature of the next path segment and for a description of how this information can be inferred from the state of distance sensors that measure the distance between the car and the track edge, see Quadflieg, Preuss, Kramer and Rudolph (2010). The network controller is also provided with two motor neurons that encode the steering and the acceleration/break command. The output of the steering neuron is normalized in the range [-1.0, 1.0]. The acceleration/brake neuron is used to accelerate and brake for values greater and lower than 0.5 with an intensity that is directly proportional to the state, in the case of acceleration, and inversely proportional to the state, in the case of deceleration.

Gears are controlled automatically by a simple hand-designed gear control that increases/decreases the current gear on the basis the car speed. Initially the gear is set to 1 and is increased of one unit when the car speed exceeds the following values [67 km/h; 114 km/h; 166 km/h; 212 km/h; 246 km/h] and decreased of one unit when the car speed decrease below the following values [232 km/h; 198 km/h; 152 km/h; 100 km/h; 53 km/h], respectively.

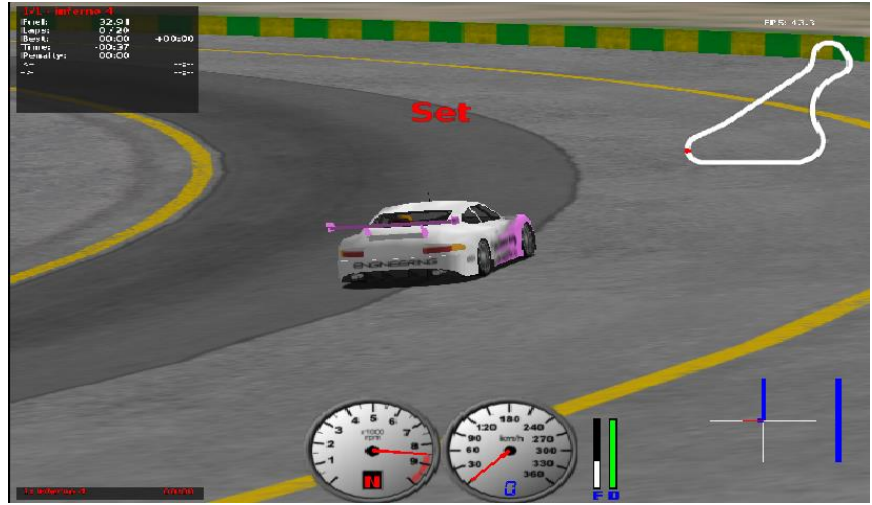

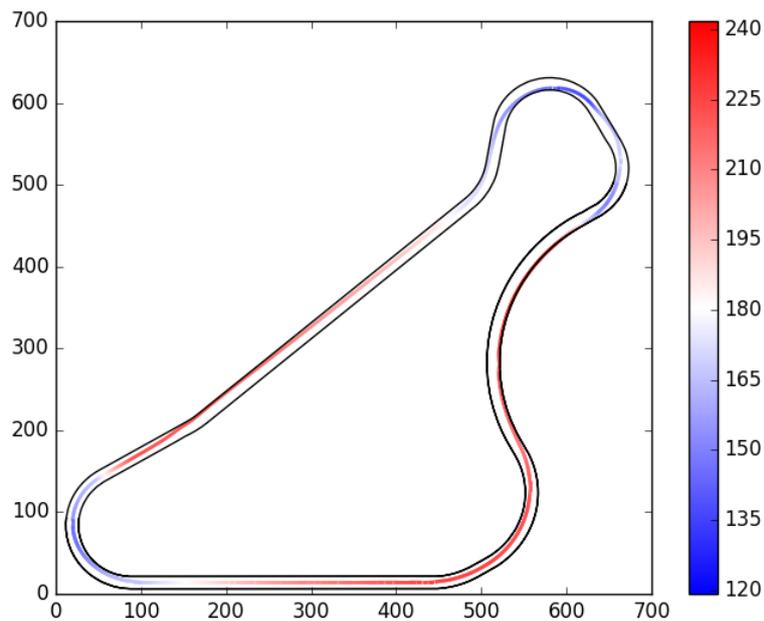

Figure 3. Top. A screenshot of the TORCS graphic render. Bottom. The CG-Track1 used in the experiments and the trajectory of the best evolved controller (see Section 3.2). The color indicates the speed in km/h during one lap of the circuit.

The experiments have been performed by using the car car2-trb1 and the CG-Track1 circuit (Figure 3). Environmental variations are introduced by setting the position of the start of the race in randomly selected locations of the circuit. Since the speed with which the car arrives in each portion of the circuit depends on the distance from the start, evolved controllers should be able to drive effectively independently of the initial speed with which they enter in each portion of the circuit.

Candidate solutions are rewarded on the basis of the distance covered during 50,000 simulation steps (which correspond approximately to 200s) and are penalized for running outside the track and for tailspins. More precisely the fitness is calculated on the basis of the following equation:

$$f = d * \left(1.0 - 0.25 * \frac{n_{out}^2}{N}\right) * (1.0 - 0.25 * a)$$

where $d$ is the distance raced in m, $n_{out}$ indicates the number of steps the car is out of the track, $N$ is the total number of steps, and $a$ is a boolean value that indicates whether or not the car produced tailspins. In the case of multiple episodes, the total fitness corresponds to the average fitness obtained during the evaluation episodes.

## 2.3 The swarm-bot foraging problem

Swarm robotics (Şahin, 2004) studies a particular class of multi-robot systems, composed of a large number of relatively simple individual robots, and emphasizes aspects like decentralization of control, robustness, flexibility and scalability. Evolutionary Swarm Robotics (Trianni, Nolfi, and Dorigo, 2008) consists in the utilization of evolutionary methods for the design of this type of systems.

Here we consider a collective foraging problem in which a group of 10 simulated marXbot robots (Bonani et al., 2010) should collect food elements and bring them to the nest (Figure 4). The robots are located in a flat square arena of 5x5 m surrounded by walls which contains a nest, i.e. a circular area with diameter of 0.8 m painted in gray. The robots, which have a circular shape and a diameter of 0.34 m, are provided with two motors controlling the desired speed of the two corresponding wheels, a ring of LEDs located around the robot body that can be turned on or off and can assume different colors, an omnidirectional camera, 36 infrared sensors located around the robot body that can detect the presence of nearby obstacles, and 8 ground infrared sensors that can be used to detect the color of the ground.

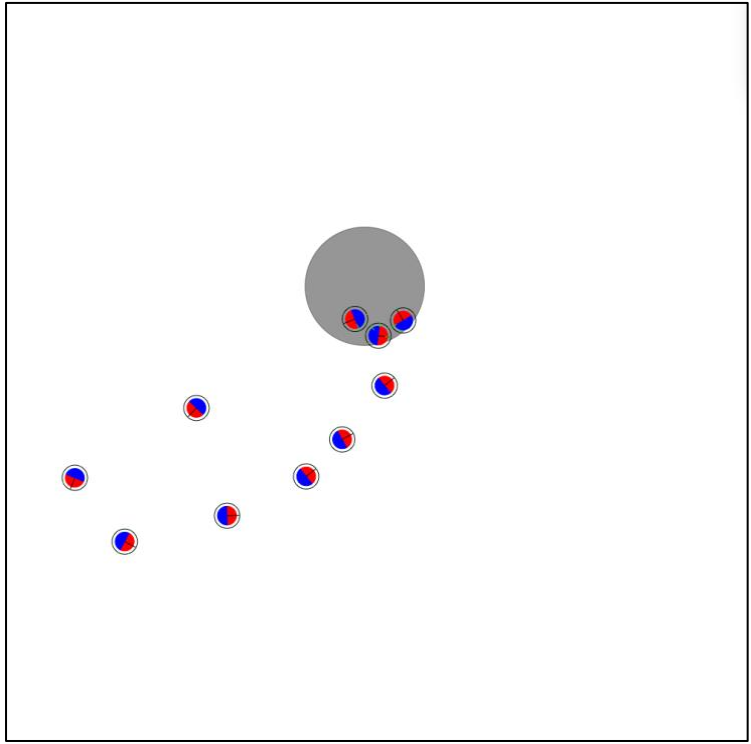

Figure 4. Top view of the environments and of the robots. The black lines and the gray circle represent the walls and the nest, respectively. The smaller circles represent the robots. The red and blue semicircles inside the robot indicate the red and blue LEDs located on the frontal and rear side of the robot.

Four hundred elements of invisible food are located inside 400 corresponding 0.25 x 0.25 m non-overlapping portions of the environment. The robots have an energy level that is replenished inside the nest and decrease linearly over time outside the nest, i.e. it is reset to 1.0 inside the nest and decreased of 0.01 every 100 ms spent outside the nest. To collect food, the energy of the robot should be greater than 0.0. Food elements are automatically collected and released when the robot enters in a portion of the environment containing a food element and in the nest, respectively. The task of the swarm consists in collecting and bringing to the nest the higher possible number of food elements within 1m and 40s. This in turn requires that the robots visit the higher possible number of environmental portions by periodically returning to the nest to release collected food elements and to replenish their energy. Fitness is calculated by simply counting the number of food elements released in the nest by the swarm.

The neural network controller of each robot is provided with the following sensory neurons: 8 units that encode the average activation of 8 groups of 4 adjacent infrared sensors, 4 units that encode the fraction of red and blue pixels detected in the frontal-left and frontal right 45º sectors of the camera, 2 units that encode the state of left and right ground infrared sensors, 1 unit that encodes the current simulated energy level, and 1 bias unit. Moreover, the controller is provided with the following motor neurons: 2 units that encode the desired speed of the robot's left and right wheels, and 2 units that binarily encode whether the 16 frontal LEDs are turned on in red or not and whether the 16 rear LEDs are turned on in blue or not.

The swarmbot is formed by identical individuals. Each candidate solution encodes the characteristics of a single neural controller that is replicated 10 times and embedded into the 10 corresponding robots.

At the beginning of each evaluation episode: (i) the nest is placed in a randomly chosen location inside the arena at a distance of at least 1m from the walls, (ii) the 10 robots are located in a randomly different positions and orientations in an area of 1x1m centered on the location of the nest, (iii) the environment is replenished with 400 food elements located in 0.25x0.25m non overlapping portions of the arena. This implies that the evolving controllers should allow the 10 robots to forage effectively irrespectively of the initial position of the nest and of the initial positions/orientations of the robots. Since the position that the robots assume during the course of the episode is influenced by their initial positions and orientations, evolved robots should be able to forage effectively in a wide range of different conditions.

In the case of this and the other two problems domains described in section 2.1 and 2.2 performance is evaluated on the basis of the best fitness of the run and on fitness distribution among multiple runs. We intentionally avoided to evaluate performance by measuring how often or how quickly near-optimum solutions are found. For a critical analysis of the latter approach see Luke (2002), Paterson and Livesey (2002) and White et al. (2013).

**2.4 The evolutionary algorithms**

To verify the possibility to evolve robust solutions for these problems and to check the relative efficacy of alternative methods in this context we carried out the experiments with five different algorithms. Four are used to evolve the connection weights of neural networks with fixed topologies. The fifth algorithm, instead, permits to evolve both the connection weights and the topology of the neural network.

The first method used consists of a standard (μ+μ) evolutionary strategy (Rechenberg, 1973; Beyer and Schwefel, 2002) that we named **stochastic steady state** (**SSS**, see Pagliuca, Milano, and Nolfi, 2018). We choose this method since it outperformed several alternative methods on two variants of the standard double-pole balancing problem in which the candidate solutions were evaluated for the ability to balance the pole from variable initial states and in which the update of sensory states was delayed of 20ms (Pagliuca, Milano, and Nolfi, 2018). It operates on the basis of populations formed by μ parents. During each generation each parent generates one offspring, the parent and the offspring are evaluated, and the best μ individuals are selected as new parents. Variations are introduced by mutating each policy parameter with a probability *mutrate* and are realized by perturbing the value with a Gaussian distribution or by replacing the value with a new random value selected with a uniform distribution. The term stochastic refers to the fact that the method includes the possibility to perturb the fitness with the addition of a randomly selected value. The parameters that should be set manually are the population size, the mutation probability, the range of the connection weights, and the range of the noise added to the fitness (when used). For a detailed description see (Pagliuca, Milano, and Nolfi, 2018).

The second method considered is the **covariance matrix adaptation evolution strategy** (**CMA-ES**, see Hansen and Ostermeier, 2001), a state-of-the-art continuous optimization method. It is a form of evolutionary strategy (ES) characterized by deterministic truncation selection, unbiased variation operators and self-adaptation, i.e. change of the behavior of variation operators during the course of the evolutionary process. Instead of an explicit population, CMA-ES maintains a multivariate Gaussian probability distribution $N(m, \sigma^2 C)$ where m and C are its mean and covariance matrix while σ is a step-size. In each iteration, the distribution is used to sample λ new candidate solutions, which are then evaluated. Afterwards, the three parameters of the distribution are updated on the basis of the observed ranking of the candidate solutions. The new mean m is calculated as a weighted recombination of the best μ candidate solutions. The update procedures of covariance matrix C and step-size σ are more complex. For a detailed description the reader is referred to the original work of Hansen and Ostermeier (2001). The population size can be set automatically on the basis of the number of parameters (Hansen and Ostermeier, 2001). Consequently, this method does not require to manually set any parameter.

The third method considered is the **exponential natural evolutionary strategy** (**xNES**, Wierstra, Schaul, Glasmachers, Sun, and Schmidhuber, 2014), a prominent member of the family of the natural evolutionary strategies (NES, see Wierstra, Schaul, Peters and Schmidhuber, 2008). NES uses the Gaussian mutation to generate new search points and adjusts the parameter of the mutation at each generation in order to improve the expected fitness under the mutation distribution. For the adjustment, the NES makes use of the natural gradient (Amari, 1998) of the expected fitness with respect to the parameter of the mutation distribution, which is referred to as a natural evolution gradient. CMA-ES and xNES are closely related and operate in a similar manner. Indeed, as demonstrated by Akimoto, Nagata, Ono and Kobayashi (2010), natural evolution strategies can be viewed as a variant of covariance matrix adaptation evolution strategies. The step-size can be set to 1.0 and the population size can be set automatically on the basis of the number of parameters. Consequently, this method does not require to manually set any parameter. For a detailed description of the method see (Wierstra, Schaul, Glasmachers, Sun, and Schmidhuber, 2014). We choose this method since it is competitive with respect to CMA-ES on black-box optimization benchmarks (Wierstra, Schaul, Glasmachers, Sun, and Schmidhuber, 2014) and since it resulted one of the best in the comparative analysis reported in (Pagliuca, Milano, and Nolfi, 2018).

The fourth method considered is the **separable natural evolutionary strategy** (**sNES**, Wierstra, Schaul, Glasmachers, Sun, and Schmidhuber, 2014), another instance of natural evolutionary strategies. The utilization of a separable and linear search distribution makes this method more suitable for problems involving high-dimensional search space (Wierstra, Schaul, Glasmachers, Sun, and Schmidhuber, 2014). For a detailed description of the method see (Wierstra, Schaul, Glasmachers, Sun, and Schmidhuber, 2014).

Finally, the fifth method considered is the **neuroevolution of augmenting topologies** (**NEAT**, Stanley and Miikkulainen, 2002), a method devised specifically for the evolution of neural network. We choose this method because it is widely used for the evolution of neural networks and permits to optimize not only the connection weights but also the topology of the network. The initial population is composed of a vector of μ genotypes of equal length that encode minimal networks in which input neurons are directly connected to output neurons. Each gene contains four numbers that encode the ID number of the neurons that send and receive the connection, the weight of the connection, and the history marker constituted by a progressive innovation number. The length of the genotype and the size of the neural network, however, can grow across generations. This occur through the usage of an add-node genetic operator, that replaces a gene encoding a connection with two genes that encode a new connection to and from a new internal neuron, and an add-connection operator that adds a new gene encoding the connection between two pre-existing neurons. The method also relies on innovation identification numbers associated to genes, which permit to cross over networks with varying topologies, and on the utilization of speciation and fitness sharing (Goldberg and Richardson, 1987) to preserve innovations. For a detailed description of the method see (Stanley and Miikkulainen, 2002). NEAT require to set many parameters. In the experiment reported below we used the parameter values recommended by the authors (Stanley and Miikkulainen, 2002).

In the case of CMA-ES, xNES and sNES the value the connection weights are unbounded and are initialized in the range [-1.0, 1.0]. In the case of the SSS and NEAT, instead, the connection weights are bounded and initialized in the range [-8.0, 8.0]. This corresponds to the value suggested by Stanley and Miikkulainen (2002) for NEAT.

The evolutionary process is continued until a total maximum number of evaluations have been performed. Since the evaluation of candidate solutions constitutes the major computational cost, this allows us to compare experiments in which the number of evaluation episodes and/or the population size differs by maintaining the computational cost approximately constant.

**2.5 Selecting solutions robust to environmental variations**

To select solutions that are robust to environmental variations we evaluate candidate solution for multiple episodes in which the environmental conditions vary stochastically.

Variations are introduced by generating an environmental variation matrix (EVM) with NEE columns and NP rows, where NEE corresponds to the number of evaluation episodes and NP to the number of environmental parameters subjected to variation. The values of the matrix are generated randomly with a uniform distribution within the range of variation of the environmental parameters ($[P1_{min}, P1_{max}]$, …. $[PN_{min}, PN_{max}]$). At the beginning of each episode the environmental conditions are set on the basis of the corresponding row of the matrix.

The number of environmental parameters subjected to variation (NP) and the variation range of each parameter is determined by the nature of the problem. The number of evaluation episodes (NEE) is set by the experimenter. Notice that the expected fitness of a candidate solution in uniformly varying environmental conditions corresponds to the average height of a fitness surface in a NP-dimensional space (as exemplified in Figure 5). Also notice how the precision of the fitness estimation depends on NEE: the higher the number of evaluation episodes is, the higher the precision of the estimation is.

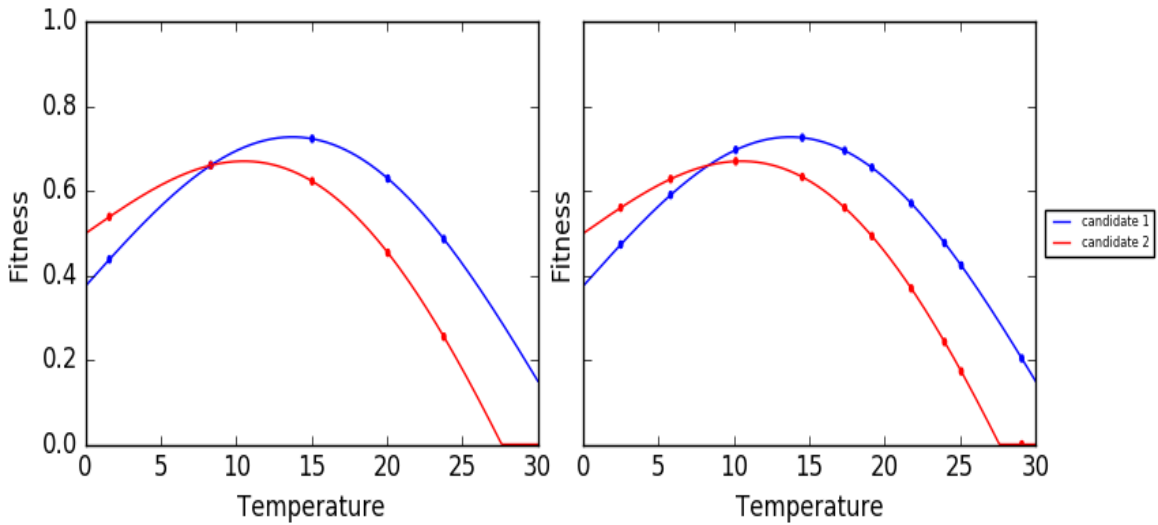

Figure 5. Illustration of the fitness surface of two hypothetical candidate solutions in an environment that varies with respect to a single parameter (e.g. an environment in which the temperature varies in the range [0, 30] ºC). Assuming that the temperature varies with an uniform distribution, the expected fitness of the two candidate solutions corresponds to the average height of the two curves, i.e. 0.55 and 0.46 for candidate solutions 1 and 2. In general, the greater the number of evaluation episodes is, the greater the precision of the estimation is. The circles in the left and right pictures show the fitness measured during 5 and 10 evaluation episodes, respectively, in which the value of the temperature was set randomly. The offset between the average value of the fitness obtained during the evaluation episodes and the expected fitness amount to 0.036 and 0.021, in the case of the left and right picture respectively.

The utilization of the same EVM for all the candidate solutions of a generation ensures that the individuals of the population are evaluated in the same conditions. This eliminates the risk of selecting the individuals that encountered easier environmental conditions to individuals with an higher expected fitness which however encountered less favorable environmental conditions.

Certain candidate solutions will result more adapted to the current environmental conditions with respect to others and consequently will have more chances to reproduce even if they are worse in other environmental conditions, especially when NEE is small. When the environmental conditions do not vary across generations, this would lead to the repeated selection of candidate solutions that are adapted to specific environmental conditions and that perform poorly in other environmental conditions. However, when the environmental conditions vary across generations, this would not happen. Indeed, the survival chances of the offspring of fragile candidate solutions, which perform well on specific environmental conditions but poorly in other environmental conditions, are low. In other words, the repeated selection in environmental conditions that vary across generations favors the evolution of candidate solutions that are robust to environmental variations even if the NEE is small, i.e. even if the number of environmental conditions in which they are evaluated is small.

The specific environmental conditions experienced during a generation create a bias that favors the selection candidate solution performing well in those specific conditions. When the environmental conditions vary across generations the direction of the bias keeps changing over generations. This continuously changing bias introduces a form of noise in the selection process that reduces the selective pressure. This effect is similar to the effect produced by the presence of noise in fitness measure, by the intentional addition of noise on the measured fitness (Jin and Branke, 2005), or by the use of stochastic selection operators such as roulette wheel selection. The presence of noise in the selection process is not necessarily negative and can even facilitates the evolution of better solutions (Bäck and Hammel, 1994; Levitan and Kauffman, 1994; Rana, Whitlev, and Cogswell, 1996), provided that the overall selection pressure is not too weak (Pagliuca, Milano, and Nolfi, 2018).

The frequency at which the environmental conditions vary across generations (F) can also influence the efficacy of the evolutionary process. Indeed, as the reported by Milano, Carvalho, and Nolfi (2017), candidate solutions evolved in environments that vary with a moderate frequency (i.e. in which the EVM matrix is regenerated every 20/50 generations) outperform candidate solutions evolved in environments that vary every generation.

The last aspect that we want to discuss is the identification of the best candidate solution of a run, or the best candidate solution evolved so far. For problems that do not need to deal with varying environmental conditions identifying the best solution is trivial: it correspond to the candidate solution that achieved the highest fitness during the run. In varying environmental conditions, instead, the solution that achieved the highest fitness tends to correspond to a solution that encountered easy environmental conditions and/or that encountered environmental conditions in which it can operate effectively. It does not necessarily correspond to the solution that produces the best performance, on the average, in all possible environmental conditions. To identify the truly best solution we might post-evaluate evolved solutions for a certain number of validation episodes (NVE) in environmental conditions that differ from those experienced by the evolving candidate solutions. This can be done by randomly generating a post-evaluation environmental matrix (PEVM) that differ from the EVM and that remains constant across generation. The higher the NVE is, the higher the accuracy of the post-evaluation is. Since post-evaluating all evolving candidate solutions is very expensive, one might restrict post-evaluation to the candidate solutions of the last generation (as in Pagliuca, Milano, and Nolfi, 2018) or to the best candidate solution of each generation, as we will do in the experiments reported in this paper.

We will use the term fitness to indicate the average score obtained during the evaluation episodes and the term performance to indicate the average score obtained during the post-evaluation episodes. Performance indicates the ability of the agents to operate in varying environmental conditions, i.e. the ability of the agents to generalize in different environmental conditions. The evolutionary process is terminated when the sum of all evaluation and post-evaluation episodes reach the total number of allowed evaluations.

## 3. Results

**3.1 The extended double-pole balancing problem**

Table 1 displays the average performance obtained in the extended double-pole balancing problem with the five evolutionary methods considered. The total number of evaluations was set to $32 * 10^6$. NEE and NVE were set to 20 and 500. Each experiment has been replicated six times by varying the frequency with which the environmental condition varied across generations. The optimal performance for this problem is unknown but it is lower than 1.0 since balancing the poles from certain initial states is likely impossible.

Table 1. Performance of the best solutions post-evaluated for 500 validation episodes for experiments in which the environmental conditions vary with different frequency across generations. F = 1 / G, where G is the number of generations after which the environmental conditions vary. Experiments conducted with NEE = 20. Each number indicate the average result of 40 replications.

|  | F=0.0 | F=0.01 | F=0.02 | F=0.04 | F=0.1 | F=1.0 |
|---|---|---|---|---|---|---|
| **SSS** | 0.043 | 0.269 | 0.313 | 0.332 | 0.335 | 0.151 |
| **CMA-ES** | 0.364 | 0.510 | 0.549 | 0.571 | 0.570 | 0.455 |
| **xNES** | 0.521 | 0.536 | 0.554 | 0.555 | 0.552 | 0.567 |
| **sNES** | 0.325 | 0.376 | 0.388 | 0.384 | 0.442 | 0.483 |
| **NEAT** | 0.174 | 0.191 | 0.194 | 0.174 | 0.165 | 0.166 |

All methods, with the exception of NEAT, achieve better performance when the environment varies across generations [0.01 >= F <= 1.0] than when the environment remains stable [F = 0.0] (Table 1, Mann-Whitney U test, p-value < 0.05 with Bonferroni correction). In the case of the CMA-ES and SSS, the best performance is achieved with intermediate frequencies: [0.02-0.1] (Mann-Whitney U test, p-value < 0.05 with Bonferroni correction for comparisons of the conditions inside the interval with the conditions outside the interval and p-value > 0.05 with Bonferroni correction for comparisons among conditions inside the interval). These results are in line with the analysis carried on the SSS by Milano, Carvalho and Nolfi (2017). In the case of xNES and sNES, instead, the best performance is achieved with high frequencies. Indeed, in the case of xNES the F=1.0 condition outperforms the F=0.0 and F=0.01 conditions (Mann-Whitney U test, p-value < 0.05 with Bonferroni correction) and does not differ statistically from the F=0.02, F=0.04, and F=0.1 conditions (Mann-Whitney U test, p-value > 0.05 with Bonferroni correction). In the case of sNES, the F=1.0 condition outperform all other conditions (Mann-Whitney U test, p-value < 0.05 with Bonferroni correction).

Table 2 displays the performance obtained by varying systematically the number of evaluation episodes from 1 to 1000. Performance is maximized in experiments carried out with the following NEE: [20, 50] SSS, [20, 50, 100, 200] CMA-ES, [50, 100, 200] xNES, [20, 50, 100, 500, and 1000] sNES, and [10, 20, 50, 100] for NEAT. The performance in the conditions indicated in the above intervals differ statistically from the other conditions (Mann-Whitney U test, p-value < 0.05 with Bonferroni correction) and do not differ statistically among themselves (Mann-Whitney U test, p-value > 0.05 with Bonferroni correction). These results indicate that the utilization of a relatively small number of evaluation episodes (i.e. 50) is sufficient to maximize performance irrespectively of the method used. Due to the tradeoff between the number of generations and number of evaluation episodes, the optimal NEE might increase and decrease, respectively, in experiments in which the total number of evaluations episode is larger or smaller. However, the fact that in these experiments the total number of evaluations episodes used is rather large suggests that the number of environmental conditions experienced by each candidate solution during its evaluation can be rather small.

Table 2. Performance of the best solutions post-evaluated for 500 validation episodes for experiments in which NEE is varied from 1 to 1000. F is set to 1.0 (i.e. the ENV matrix is re-generated every generation). Each number indicate the average result of 40 replications.

|  | NEE=1 | NEE=10 | NEE=20 | NEE=50 | NEE-100 | NEE=200 | NEE=500 | NEE=1000 |
|---|---|---|---|---|---|---|---|---|
| **SSS** | 0.099 | 0.213 | 0.151 | 0.123 | 0.091 | 0.085 | 0.082 | 0.088 |
| **CMA-ES** | 0.125 | 0.413 | 0.455 | 0.477 | 0.483 | 0.504 | 0.441 | 0.412 |
| **xNES** | 0.087 | 0.508 | 0.567 | 0.605 | 0.607 | 0.600 | 0.558 | 0.517 |
| **sNES** | 0.139 | 0.421 | 0.483 | 0.500 | 0.507 | 0.484 | 0.475 | 0.465 |
| **NEAT** | 0.068 | 0.153 | 0.166 | 0.150 | 0.138 | 0.120 | 0.083 | 0.064 |

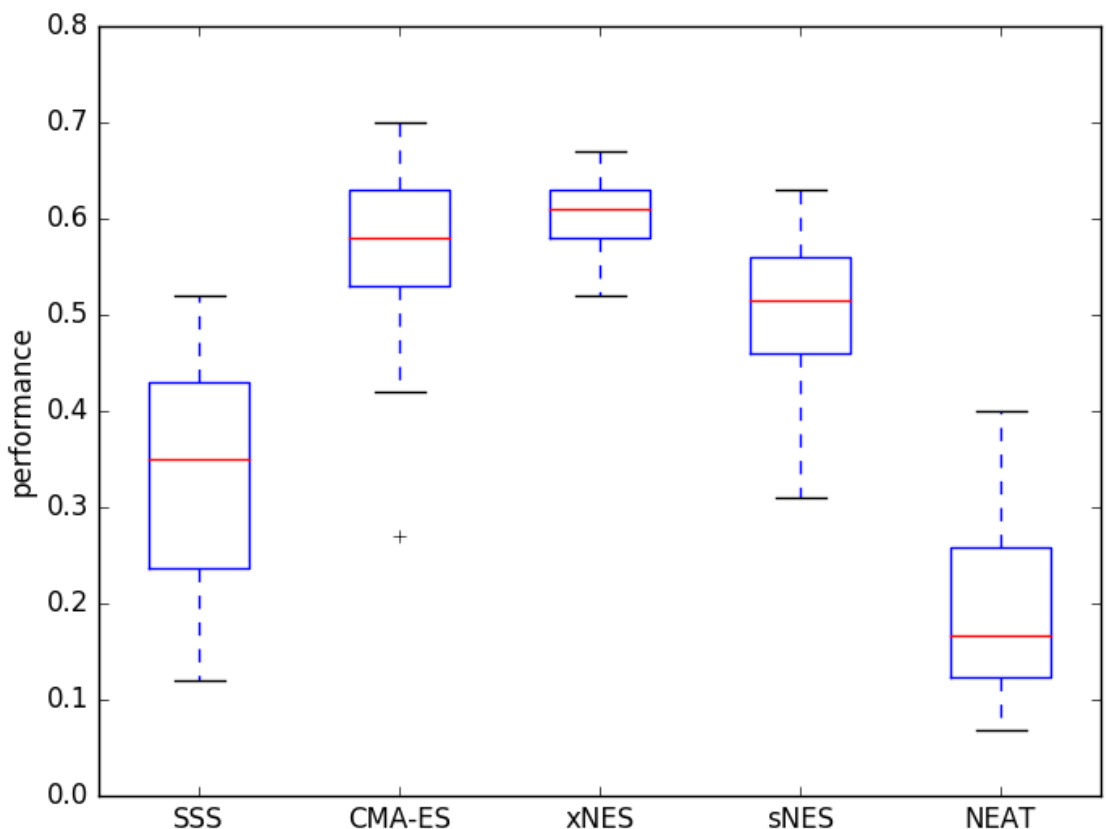

Figure 6. Performance of the best candidate solutions post-evaluated for 500 trials. Data refer to the best combination of parameters for each algorithm: SSS (F = 0.1, NEE = 20); CMA-ES (F = 0.04, NEE = 20); xNES (F = 1.0; NEE = 100); sNES (F = 1.0, NEE = 100); NEAT (F = 0.02, NEE = 20). Boxes represent the inter-quartile range of the data and horizontal lines inside the boxes mark the median values. The whiskers

extend to the most extreme data points within 1.5 times the inter-quartile range from the box. "+" indicate the outliers. Data obtained by running 40 replications.

Figure 6 displays the performance obtained with the five methods with the best combination of parameters. The best performance is achieved by CMA-ES and xNES that outperform the other three methods (Mann-Whitney U test, p-value < 0.05 with Bonferroni correction) and do not statistically differ among themselves (Mann-Whitney U test, p-value > 0.05 with Bonferroni correction). The second in the ranking is the sNES method that is outperformed by CMA-ES and xNES (Mann-Whitney U test, p-value < 0.05 with Bonferroni correction) and outperforms SSS and NEAT (Mann-Whitney U test, p-value < 0.05 with Bonferroni correction). The third in the ranking are the SSS and NEAT methods that are outperformed by CMA-ES, xNES, and sNES (Mann-Whitney U test, p-value < 0.05 with Bonferroni correction) and do not differ among themselves (Mann-Whitney U test, p-value > 0.05 with Bonferroni correction).

The analysis of the best-to-date performance across generations shows how xNES, CMA-ES, and sNES outperform the other two methods from the very beginning for the entire course of the evolutionary process (Figure 7).

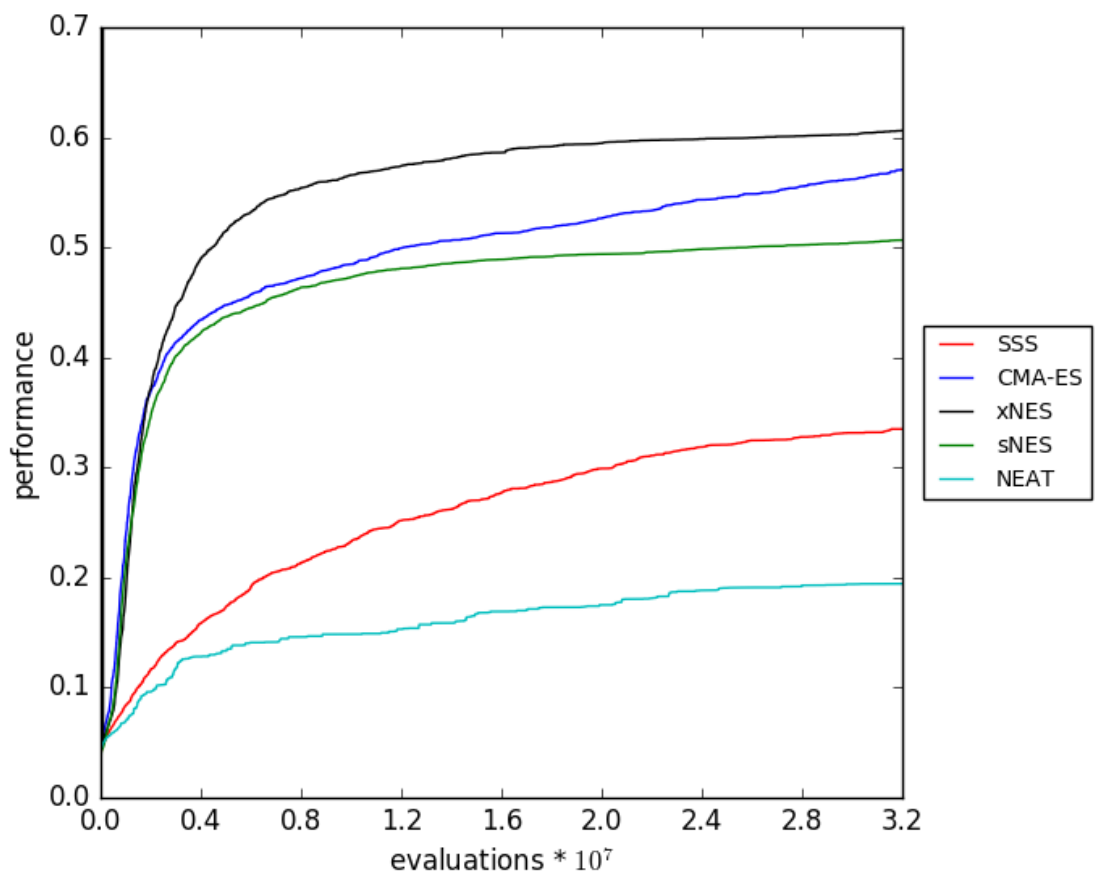

Figure 7. Performance of the best candidate solution so-far across generations for different methods. The X-axis indicate evaluations. Data obtained in the experiments with the best combination of parameters (see caption of Figure 6). Each curve indicates the average results of 40 replications.

Figure 8 shows the percentage of times in which the best solution of a run is generated in one of the 10 consecutive phase of the evolutionary process. As expected, in the majority of the cases the best solution of a run is generated during the final phases of the evolutionary process (with the exception of NEAT). However, in a non-negligible fraction of the runs, the best solution is generated during intermediate and even during initial phases of the evolutionary process. This indicate that post-evaluating the best candidate solution of each generation is necessary to identify the best solution of a run.

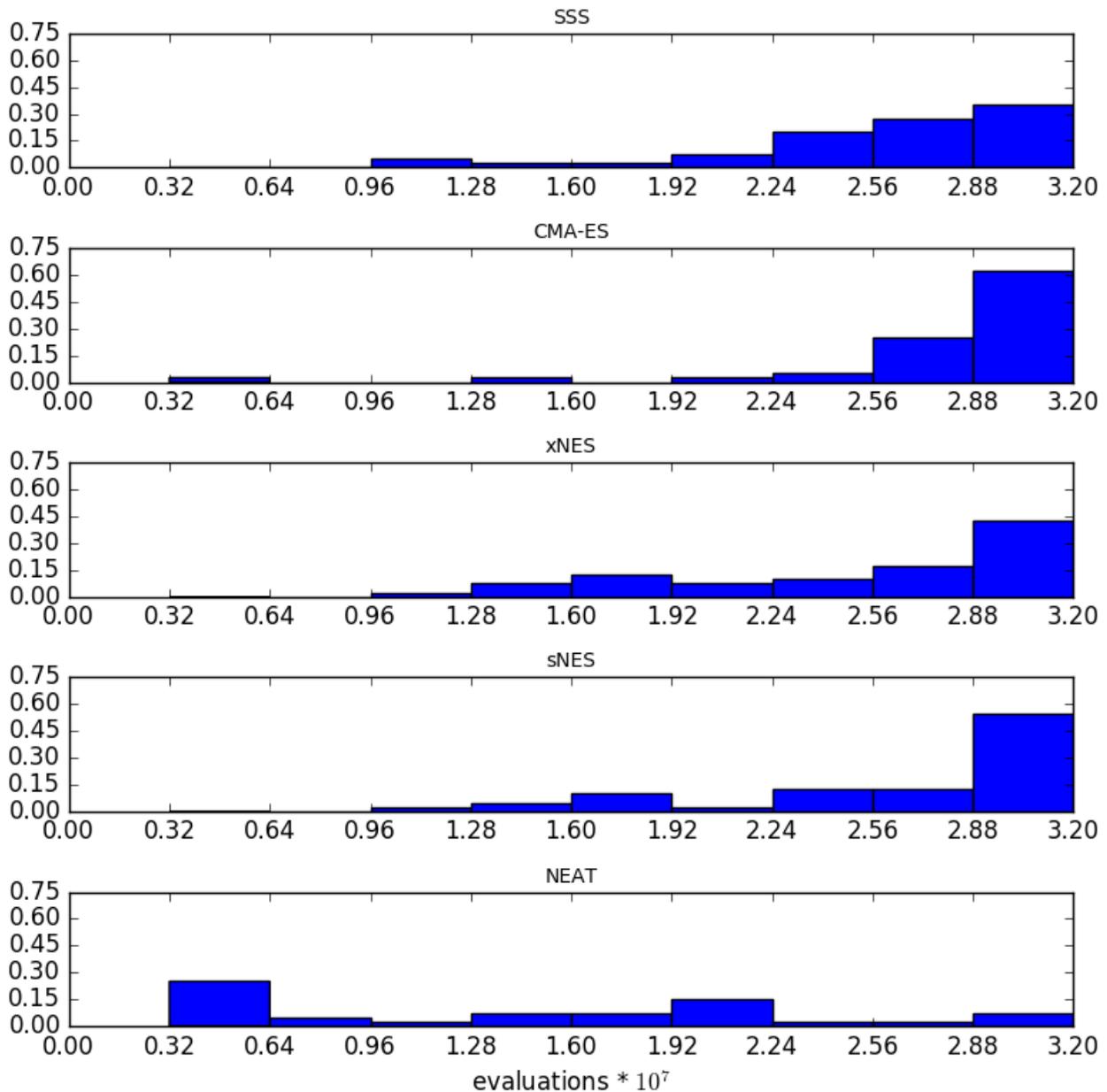

Figure 8. Fraction of times in which the best solution was generated during one of ten consecutive phases of the evolutionary process. Each phase corresponds to $32 * 10^5$ evaluations for a total of $32 * 10^6$ evaluations. Data computed on the experiments performed with the SSS, CMAES, xNES, sNES, and NEAT methods with the best combination of parameters (see Figure 6). Each histogram shows the fraction of run, out of 40, in which the best solution was found during the corresponding phase.

### 3.2 The car-racing problem

Figure 9 displays the performance obtained in the car racing problem with the five evolutionary methods considered. The total number of evaluations was set to $1 * 10^6$. NEE and NVE were set to 3 and 25, respectively. The frequency of environmental variation (F) was set to 0.05, i.e. to a value that produced good performance with the different methods, on the average, in the case of the extended double-poles problem. Each experiment has been replicated 20 times.

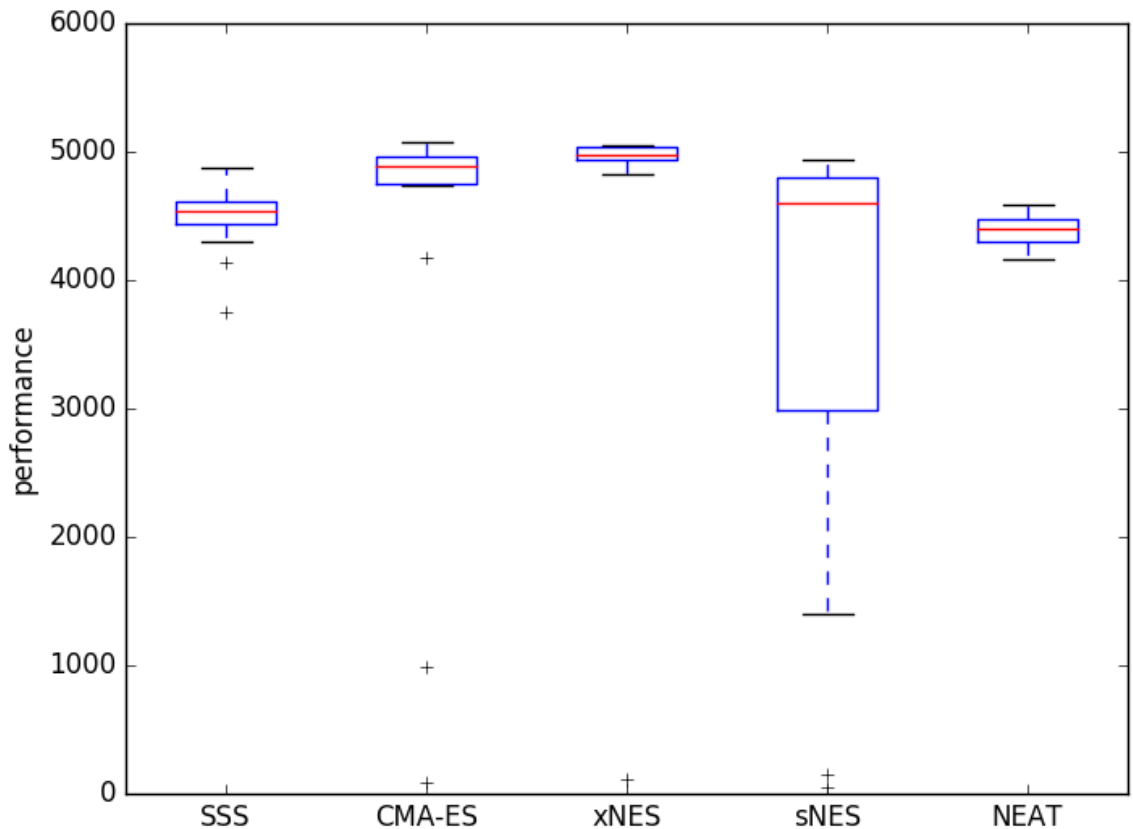

Figure 9. Performance of the best candidate solutions post-evaluated for 25 trials. Each experiment has been replicated 20 times. Boxes represent the inter-quartile range of the data and horizontal lines inside the boxes mark the median values. The whiskers extend to the most extreme data points within 1.5 times the inter-quartile range from the box. "+" indicate the outliers.

Also in this case the best performance is achieved with the xNES and CMA-ES methods that outperform the sNES, SSS, and NEAT methods (Mann-Whitney U test, p-value < 0.05 with Bonferroni correction) and do not differ among themselves (Mann-Whitney U test, p-value > 0.05 with Bonferroni correction). The performance of the sNES, SSS and NEAT methods do not statistically differ among themselves (Mann-Whitney U test, p-value > 0.05 with Bonferroni correction).

The optimal performance for this problem is unknown. Remarkably, however, the best controllers obtained with the CMA-ES, sNES, xNES and SSS methods outperform the best human-designed controller named "Inferno" included in the Torque repository (Table 3).

Table 3. Performance of the best controllers evolved with the 5 methods and of the best human-designed driver available in TORCS v1.3.7 called "Inferno" (i.e. the human programmed controller that achieves the best performance on the CG1-track circuit, see

http://torcs.sourceforge.net/index.php?name=News&file=article&sid=100). All controllers, including Inferno, have been tested and evaluated in the same conditions.

| CMA-ES | xNES | sNES | SSS | Inferno | NEAT |
|---|---|---|---|---|---|
| 5078.018 | 5047.128 | 4918.661 | 4860.633 | 4658.589 | 4581.371 |

Inferno has access to the full description of the circuit and operates by initially computing an optimal racing line. The program then determines the state of the steering, the acceleration and break actuators of the car at every time-step on the basis of the offset between the current state of the car and the future desired state of the car along the optimal racing line and on the basis of the current speed of the car. The way in which the state of the actuators of the car are controlled is also regulated on the basis of phase of the race (starting versus normal phase) and on the basis of the risk of penalties (dangerous versus non-dangerous context). Gears are controlled on the basis of the same routine used by evolving controllers and described in Section 3.2.

Also in the case of this problem, the utilization of a limited number of evaluation episodes is sufficient to evolve controller able to generalize with respect to the initial position of the car in the circuit.

### 3.3 The swarmbot foraging problem

Figure 10 displays the performance obtained on the swarmbot foraging problem with the five evolutionary methods considered. The total number of evaluations was set to $1.5 * 10^6$. NEE and NVE were set to 6 and 20, respectively. The frequency of environmental variation (F) was set to 0.05, i.e. to a value that produced good performance with the different methods, on the average, in the case of the extended double-pole problem.

In this problem the best performance is achieved by the CMA-ES, xNES, and sNES methods that outperform the SSS and NEAT methods (Mann-Whitney U test, p-value > 0.05 with Bonferroni correction) and do not statistically differ among themselves (Mann-Whitney U test, p-value > 0.05 with Bonferroni correction). The SSS and NEAT methods do not statistically differ among themselves (Mann-Whitney U test, p-value > 0.05 with Bonferroni correction).

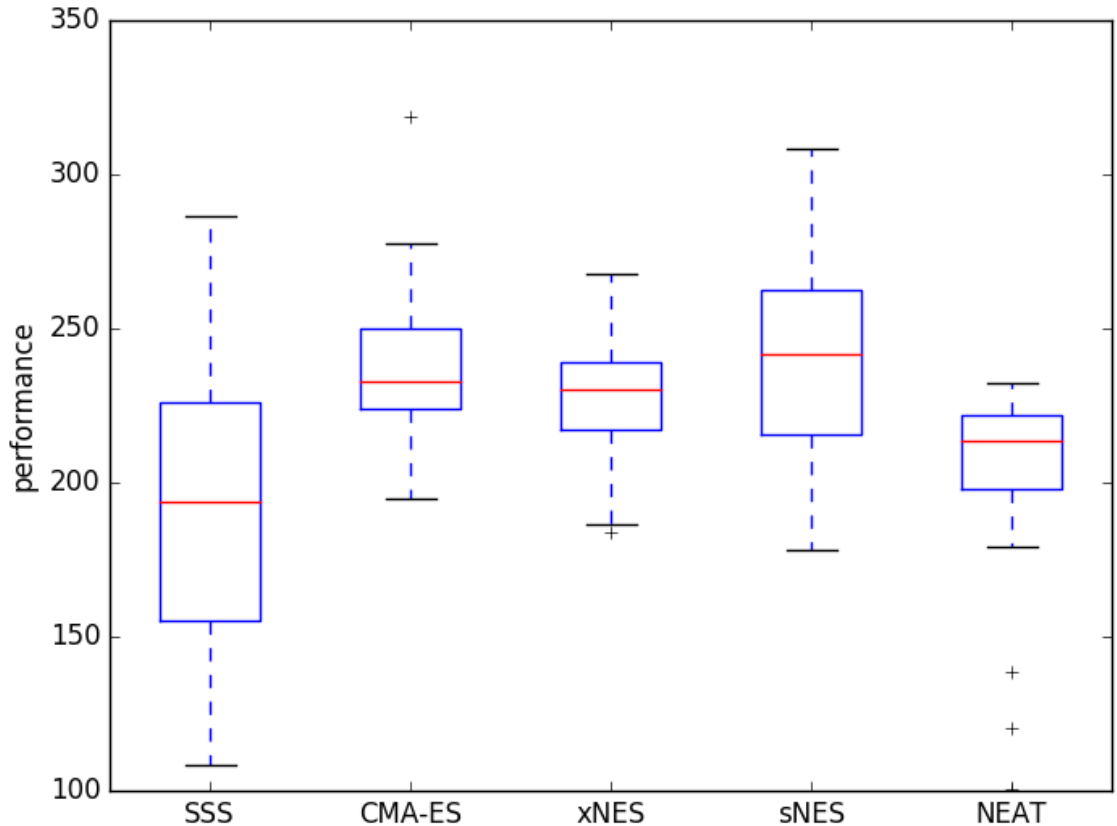

Figure 10. Performance of the best solution post-evaluated for 20 episodes. Boxes represent the inter-quartile range of the data and horizontal lines inside the boxes mark the median values. The whiskers extend to the most extreme data points within 1.5 times the inter-quartile range from the box. "+" indicate the outliers. Data have been obtained by running 30 replications.

The analysis of the behavior displayed by the best evolved swarmbots indicates that evolution manages to discover rather effective and elaborated solutions. The most common and effective solution type involves the formation of a dynamic chain in which the robots arrange themselves in an ellipsoid shape (Figure 4) and move at high speed approximately toward the position of the next robot so to maintain the ellipsoid formation. One side of the ellipsoid chain is located over the nest so to allow the robots to periodically release the collected food items in the nest. The location of the other side of the chain tends to move slowly over time so to permit the swarmbot to collect food elements located in different areas of the environment (see video S1). A common strategy that the robots use to ensure that one side of the ellipsoid remains anchored on the nest consists in slowing down when they enter in the nest and exiting from the nest only when they detect another robot nearby on their rear side (see video S2). A common strategy that is used to ensure that the second side of the ellipsoid varies its position consists in re-entering in the nest after a complete lap of the ellipsoid on the left or on the right of the previous time. Indeed, this ensures that the second side of the ellipsoid slowly moves clockwise or counter-clockwise over time (see video S1). Robots relying on this type of strategy turn their rear blue red on and use the perception of blue to form and to move along the ellipse of the dynamic chain.

    A second type of strategy involve the subdivision of the swarm in two sub-groups of robots: a larger group that collect food and a smaller group (or eventually a single robot) that mark the position of the nest (see video S3). The larger group of food collectors is constituted by individuals

that move independently in different directions collecting food. The smaller group of nest-marker act as a beacon and enable the former robots to infer the position of the nest by distance and to navigate back to the nest to release the collected food and replenish the energy level. The role of the individual robots might change dynamically since food-collectors returning to the nest can become nest-markers and since nest-markers might exit from the nest and become food-collectors. The transition between the two roles is regulated so to ensure that the number of robots playing the role of nest-marker is smaller and eventually is formed by a single individual.

The behavior of evolved robots is rather robust. They manage to: (i) form the dynamic chains or to subdivide in food-collectors and nest-markers during all trials, (ii) avoid colliding with obstacles and with other moving robots during the large majority of the evaluation episodes, and (iii) collect and to bring to the nest a significant number of food elements in all trials, irrespectively of their initial positions and orientations and irrespectively of the relative positions and orientations of the other individuals. In the case of the robots displaying the dynamic chain strategy, occasionally part of the robots might exit from the chain and form a sub-chain than usually re-merge with the main chain.

**4. Conclusions**

We investigated the possibility to evolve solutions that are robust with respect to variations of the environmental conditions and that consequently operate effectively in new conditions immediately, without the need to adapt to variations. After reviewing related research and discussing the open research issues we proposed a suitable method that specify how the fitness of candidate solutions can be estimated, how the environmental condition should vary to facilitate the evolution of effective solutions, and how the best solution of a run can be identified. Moreover, we analyzed the characteristics that makes evolutionary algorithms suitable for the evolution of solutions robust to environmental variations.

The method is characterized by the following aspects: (i) candidate solutions are evaluated multiple times in a limited number of randomly different environmental conditions, (ii) the candidate solutions of a given generation are evaluated in the same environmental conditions, (iii) the environmental conditions change across generations with a given frequency, (iv) the best solution of a run is identified by post-evaluating the best candidate solution of each generation on a independent set of randomly different environmental conditions. The identification of algorithms that are suitable to evolve robust solution has been made by comparing the efficacy of five different state-of-the-art evolutionary algorithms for the evolution of robust solutions with the method described above.

The results of the comparison indicate that algorithms that optimize a distribution of parameters (i.e. CMA-ES, xNES, and sNES) outperform algorithms that optimize a particular vector of parameters (i.e. SSS and NEAT). The advantage of the former methods can be due to their ability to estimate the gradient of the fitness and/or to their ability to select solutions that are robust with respect to variations of their parameter (see Lehman, Chen, Clune and Stanley, 2018). This since robustness to parameter variation might correlate positively with robustness to variation of the environmental conditions (Lehner, 2010). The relative importance of gradient estimation and of robustness to variations of parameters for the synthesis of solutions robust to environmental variations should be investigated in future research.

More specifically the CMA-ES and xNES algorithms outperform the SSS and NEAT methods in all considered problem domains. The sNES algorithm work as well as the CMA-ES and xNES methods in one problem but produces lower performance in the other two problems. Variations of these methods specially tailored for neuro-evolution should be explored in future research. In particular, we believe that the possibility to increase the size of the neural network across generations through a mechanism analogous to that used by NEAT (Stanley and Miikkulainen, 2002) and weight decay mechanisms that prevent an excessive growth of parameters represent two promising research directions.

The results obtained on three qualitatively different problems involving a relatively large number of parameters to be optimized show how the method proposed is effective and computational tractable, i.e. permits to evolve solutions that operate effectively in a wide range of environmental conditions and rely on a limited number of evaluation episodes. It allows to: (i) achieve better performance on the extended double-poles balancing problem with respect to previous research, (ii) to evolve controllers that outperform the best available human-designed controller in the case of the car-racing problem, and (iii) to synthesize remarkable effective solutions in the case of the swarmbot foraging problem (see Section 3.1, 3.2 and 3.3).

**Additional online material**

Video S1. Available from http://laral.istc.cnr.it/res/robust-optimization/video1.mov
Video S2. Available from http://laral.istc.cnr.it/res/robust-optimization/video2.mov
Video S3. Available from http://laral.istc.cnr.it/res/robust-optimization/video3.mov

**Acknowledgment**

The authors thanks Jonata Carvalho and Nicola Milano for useful comments.